\documentclass[final]{cvpr}

\usepackage{times}
\usepackage{epsfig}
\usepackage{listings}
\usepackage{graphicx}
\usepackage{amsmath}
\usepackage{amssymb}
\usepackage{dirtree}
\usepackage{diagbox}
\usepackage{booktabs}
\usepackage{subcaption}

\usepackage[pagebackref=true,breaklinks=true,colorlinks,bookmarks=false]{hyperref}

\def\cvprPaperID{5} 
\def\confYear{CVPR 2021}

\begin{document}

\title{Multi-resolution Outlier Pooling for Sorghum Classification}

\author{Chao Ren$^1$, Justin Dulay$^1$, Gregory Rolwes$^1$, Duke Pauli$^2$, Nadia Shakoor$^3$, and Abby Stylianou$^1$ \\
$^1$ Saint Louis University\\
$^2$ University of Arizona\\
$^3$ Donald Danforth Plant Science Center
}

\maketitle

\begin{abstract}
Automated high throughput plant phenotyping involves leveraging sensors, such as RGB, thermal and hyperspectral cameras (among others), to make large scale and rapid measurements of the physical properties of plants for the purpose of better understanding the difference between crops and facilitating rapid plant breeding programs. One of the most basic phenotyping tasks is to determine the cultivar, or species, in a particular sensor product. This simple phenotype can be used to detect errors in planting and to learn the most differentiating features between cultivars. It is also a challenging visual recognition task, as a large number of highly related crops are grown simultaneously, leading to a classification problem with low inter-class variance. In this paper, we introduce the Sorghum-100 dataset, a large dataset of RGB imagery of sorghum captured by a state-of-the-art gantry system, a multi-resolution network architecture that learns both global and fine-grained features on the crops, and a new global pooling strategy called Dynamic Outlier Pooling which outperforms standard global pooling strategies on this task.
\end{abstract}

\begin{figure}[ht!]
    \centering
    \includegraphics[width=.85\columnwidth]{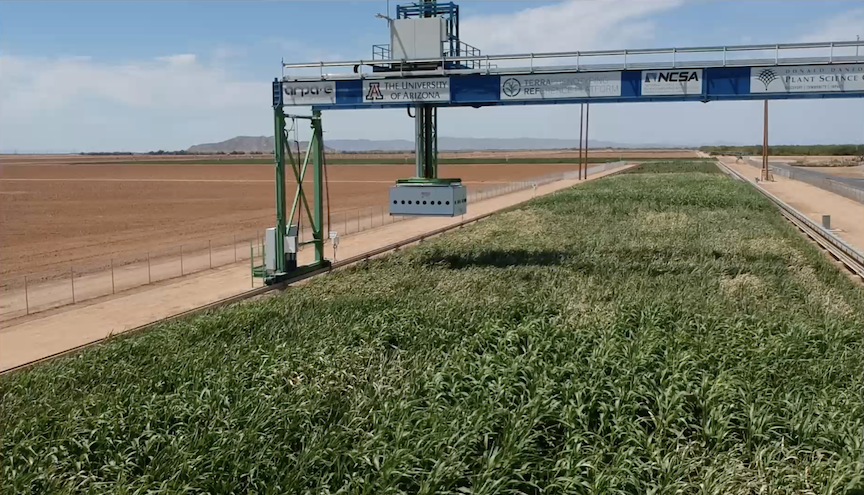}
    \includegraphics[width=.85\columnwidth]{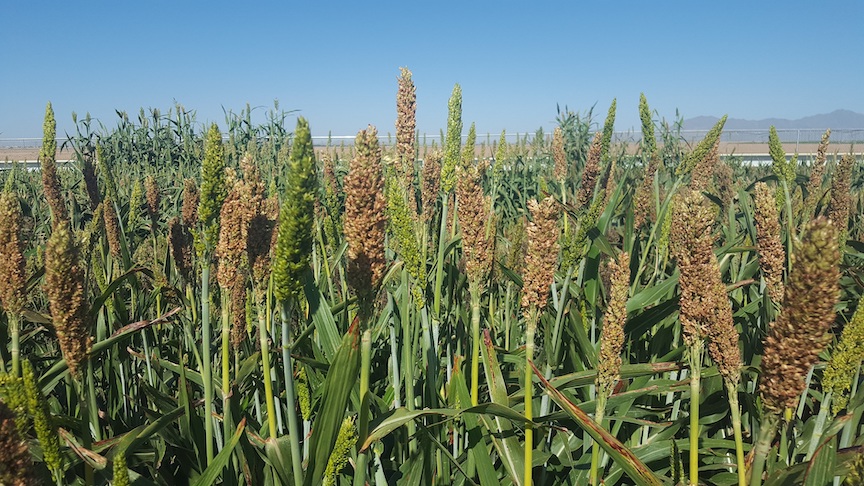}
    \includegraphics[width=.85\columnwidth]{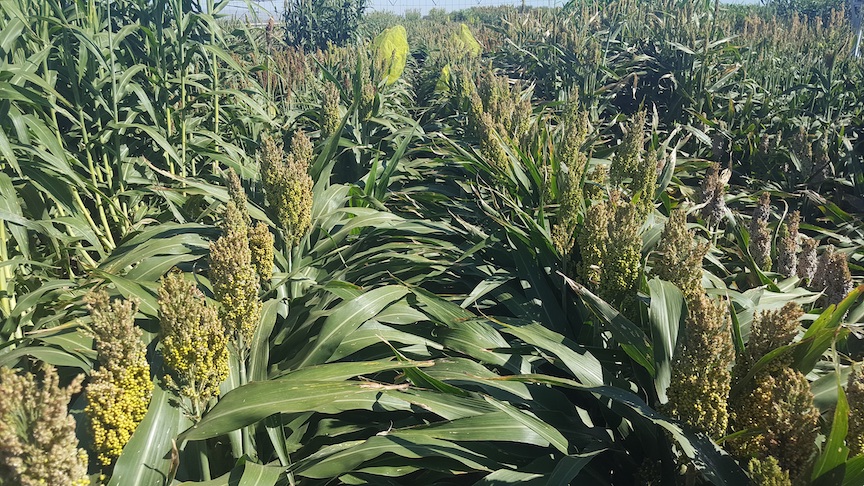}
    \caption[TERRA-REF Field Scanner]{The TERRA-REF Field and Gantry-based Field Scanner in Maricopa, Arizona~(top), with sorghum being grown in the field. Sorghum is a hugely important cereal crop, widely used as a source of grain, agricultural feed, and even bio-fuel. Over several seasons, hundreds of varieties of both bio-energy and grain sorghum were grown in the TERRA-REF field (middle and bottom), and were imaged daily for the purpose of high throughput phenotyping.}
    \label{fig:field_scanner}
\end{figure}

\section{Introduction}
Sorghum is widely used as an agricultural feed substitute, a gluten-free ancient grain, a source of bio-fuel, and even as popcorn in some food communities. Demand for sorghum for a variety of purposes has risen with the need for better food and energy sources, motivating the need to rigorous plant breeding strategies to select for traits that are valued for each purpose (e.g., more grain for food uses or more biomass for bio-fuel production).

Automated high throughput plant phenotyping involves leveraging sensors, such as RGB, thermal and hyperspectral cameras (among others), to make large scale and rapid measurements of the physical properties of plants for the purpose of better understanding the difference between crops and facilitating rapid plant breeding programs. One of the most basic phenotyping tasks is to determine the cultivar (or species) in a particular sensor product. In experiments with a large number of related cultivars being grown simultaneously, this is a challenging fine-grained visual categorization task due to the low inter-class variability.

Our contributions towards this high throughput phenotyping task are threefold:
\begin{itemize}
    \item A large scale sorghum cultivar classification dataset consisting of tens of thousands of images captured from a gantry based RGB sensor;
    \item A multi-resolution network architecture that learns relevant features at different scales; and
    \item A new global pooling strategy called Outlier Pooling which significantly outperforms standard pooling approaches in this agricultural setting.
\end{itemize}

\section{Background}

\subsection{TERRA-REF Field and Gantry-based Field Scanner}
\begin{figure}
    \centering
    \includegraphics[width=.49\columnwidth]{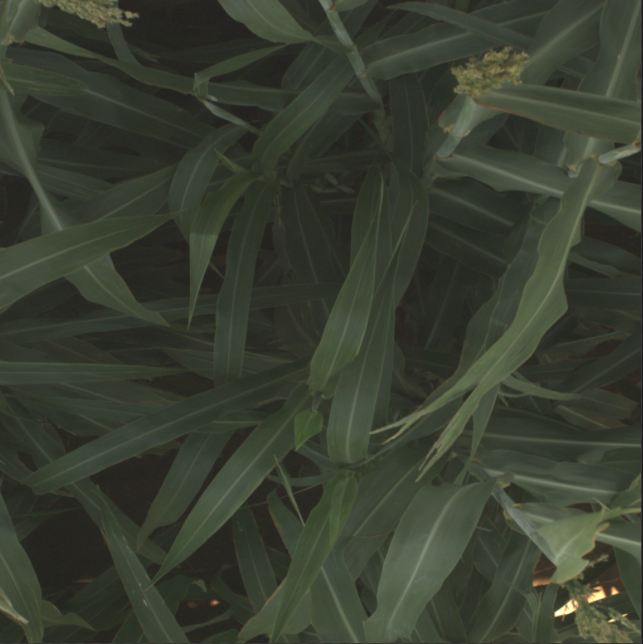}
    \includegraphics[width=.49\columnwidth]{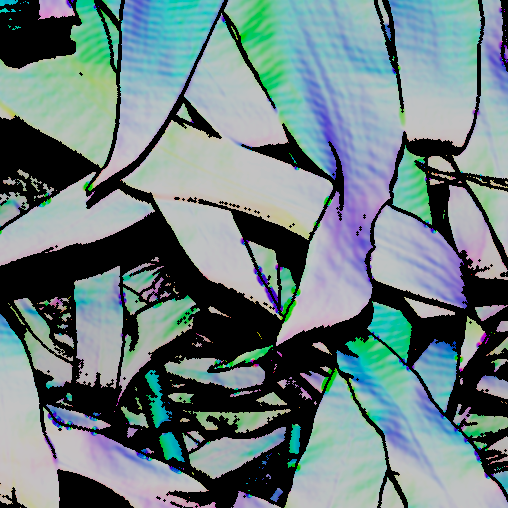}
    \includegraphics[width=\columnwidth]{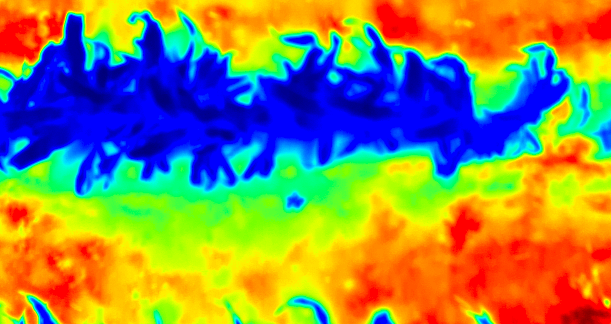}
    \caption[Feature Extraction]{\textbf{Example data from the TERRA-REF gantry system.} (top-left) RGB data. (top-right) 3D-scanner data (false color, where color indicates the surface normal and value indicates depth from the scanner).~(bottom) Thermal data. In this paper we focus on data from the RGB camera.}
    \label{fig:example_data}
\end{figure}

In 2016, the Advanced Research Project Agency--Energy (ARPA-E) funded the Transportation Energy Resources from Renewable Agriculture Phenotyping Reference Platform, or TERRA-REF\cite{terra,LeBauer2020}. The TERRA-REF project stood up a state-of-the-art gantry based system for monitoring the full growth cycle of over an acre of crops with a cutting-edge suite of imaging sensors, including stereo-RGB, thermal, short- and long-wave hyperspectral, and laser 3D-scanner sensors. The goal of the TERRA-REF gantry was to perform in-field automated high throughput plant phenotyping, the process of making phenotypic measurements of the physical properties of plants at large scale and with high temporal resolution, for the purpose of better understanding the difference between crops and facilitating rapid plant breeding programs. Due to the technical demands of high-throughput phenotyping, it is most often performed in controlled environments (e.g., greenhouses with imaging platforms). Controlled  environments  play  a  very  important  role  in  understanding  plant performance by providing management of the abiotic environment, and the ability to reproduce experimental conditions year-round, but plant performance in field settings, both in terms of growth and yield parameters, is strongly influenced by variability in weather, soil conditions and other environmental parameters that cannot be observed in the greenhouse. The TERRA-REF gantry system was designed to meet the sort of technical requirements for high throughput phenotyping in a field setting. The TERRA-REF field and gantry system are shown in Figure~\ref{fig:field_scanner}, and example data captured from its RGB, 3D-scanner and thermal cameras are shown in Figure~\ref{fig:example_data}.

Over the course of its first several years in operation, the TERRA-REF platform collected multiple petabytes of sensor data capturing the full growing cycle of sorghum plants from the sorghum Bioenergy Association Panel~\cite{brenton2016genomic}, a set of 390 sorghum cultivars whose genomes have been fully sequenced and which show promise for bio-energy usage.

The full, original TERRA-REF dataset in its entirety is a massive public domain agricultural dataset, with high spatial and temporal resolution across numerous sensors and seasons, and includes a variety of environmental data and extracted phenotypes in addition to the sensor data. More information about the dataset and access to is can be found in~\cite{LeBauer2020}.

\subsection{Phenotyping from Aerial Data}
While in-field sorghum phenotypes have been determined using aerial RGB data from drones\cite{ribera2018estimating,masjedi2018sorghum,chen2017plant,zhang2017_biomass,guo2018_sorghum,potgieter2017_uav}, UAV datasets are limited in their temporal resolution due to the labor required in capturing the data, and their spatial resolution is limited both by the sensors that are able to be mounted on board a drone and the time constraints of the drone operator (flights from lower to the ground are higher resolution but take longer to complete). By comparison, the data from the TERRA-REF Gantry-based Field Scanner has both high spatial resolution (the gantry has extremely high quality sensors and the height of the gantry is placed to optimally image the plants thoughout the growing cycle), and temporal resolution (data is captured every day).



\section{Dataset \& Classification Task}
\begin{figure*}
    \centering
    \includegraphics[width=\textwidth]{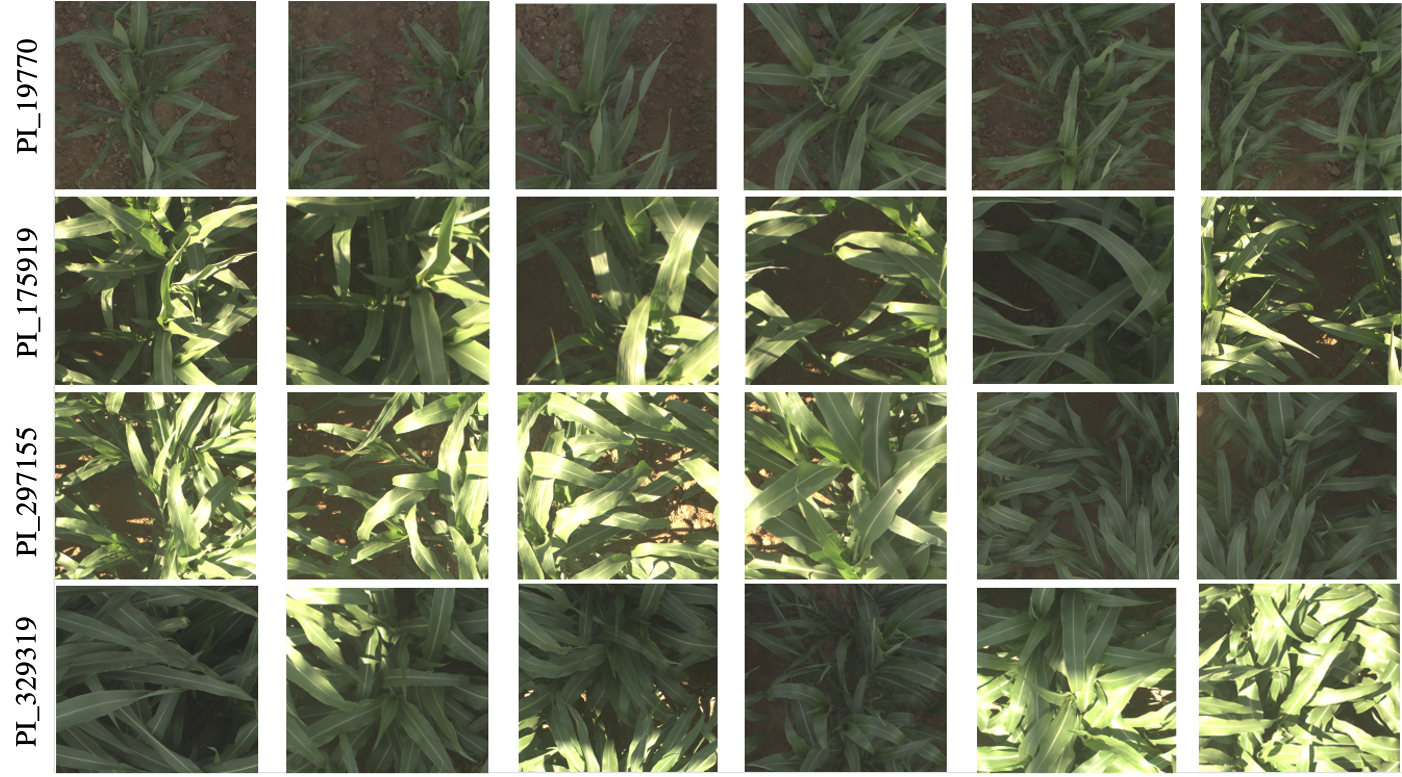}
    \caption[Large Crops Comparison]{Multiple images from the dataset representing different cultivars. The rows represent different culitvars. The columns represent different captured dates respectively: June 1st, 3rd, 7th, 17th, 19th and 27th, 2017.}
    \label{fig:4x6_cultivars}
\end{figure*}

In this paper, we describe the Sorghum-100 dataset, a curated subset of the RGB imagery captured during the TERRA-REF experiments, labeled by cultivar and day after planting. The dataset will be released publicly and there will be a corresponding Kaggle competition. This data could be used to develop and assess a variety of plant phenotyping models which seek to answer questions relating to the presence or absence of desirable traits (e.g., ``does this plant exhibit signs of water stress?''). In this paper we focus on the question: ``What cultivar is shown in this image?'' Predicting the cultivar in an image is an especially good challenge problem for familiarizing the machine learning community with the TERRA-REF data. At first blush, the task of predicting the cultivar from an image of a plant may not seem to be the most biologically compelling question to answer -- in the context of plant breeding, the cultivar, or parental lines are typically known. A high accuracy machine learning predictor of the species captured by the sensor data, however, can be used to determine where errors in the planting process may have occurred. For example, seed may be mislabeled prior to planting, or planters may get jammed, depositing seeds non-uniformly in a field~\cite{sharma2019comparison}. Both types of errors are surprisingly common and can cause major problems when processing data from large-scale field experiments with hundreds of cultivars and complex field planting layouts.

The Sorghum-100 dataset consists of 48,106 images and 100 different sorghum cultivars grown in June of 2017 (the images come from the middle of the growing season when the plants were quite large but not yet lodging -- or falling over). In Figure~\ref{fig:4x6_cultivars}, we show a sample of images from four different cultivars. Each row includes six images from different dates in June. This figure highlights the high inter-class visual similarity between the different classes, as well as the high variety in the imaging conditions from one day to the next, or even over the course of a day.

\begin{figure}
    \centering
    \includegraphics[width=\columnwidth]{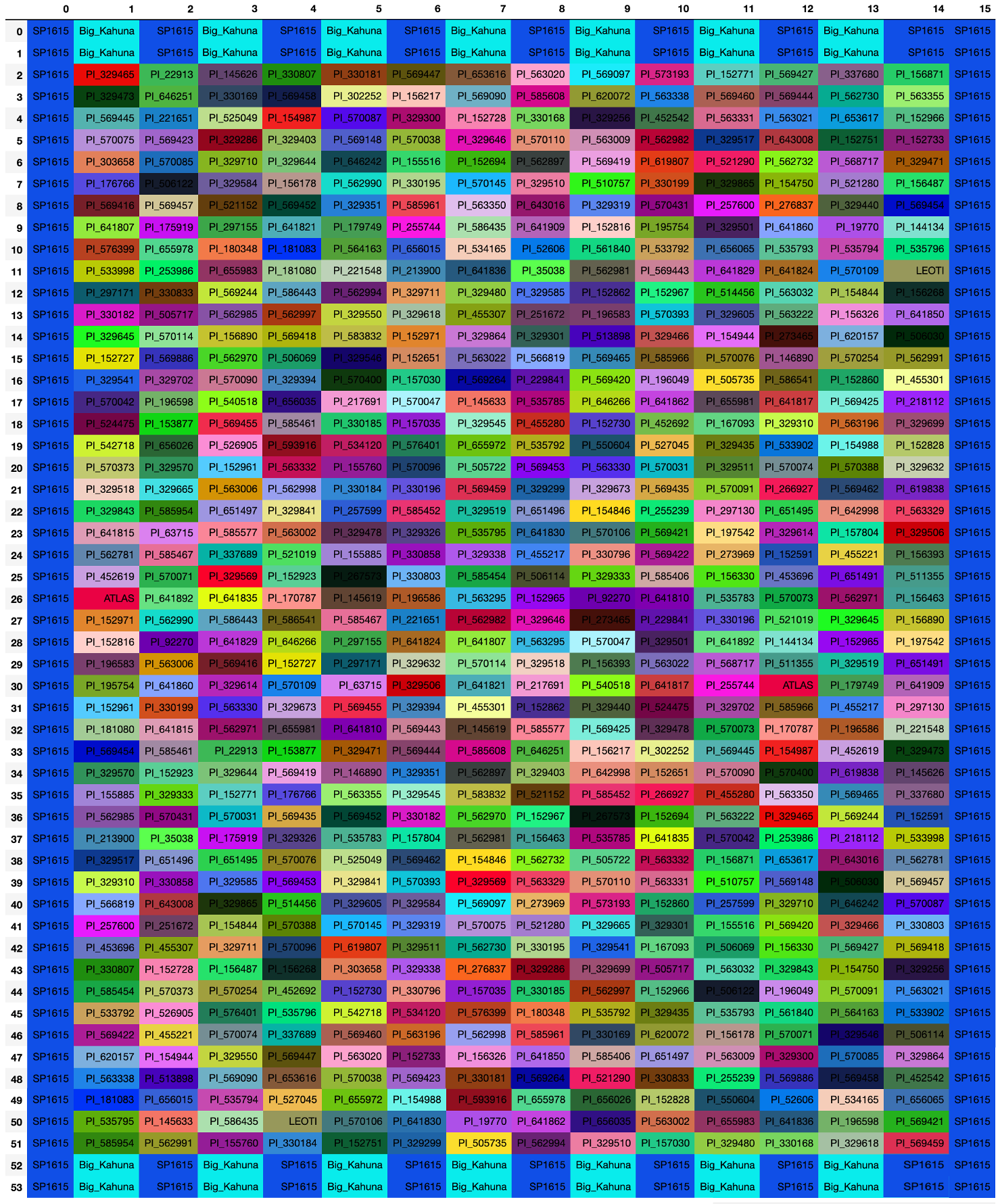}
     \includegraphics[width=\columnwidth]{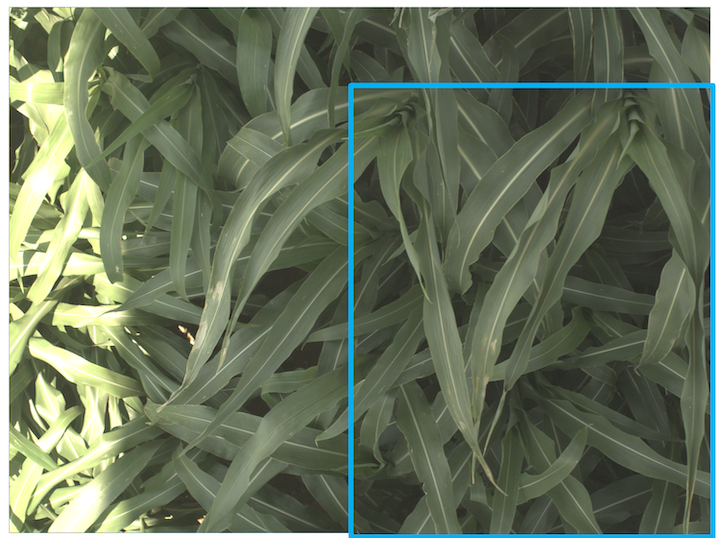}
    \caption{\textbf{TERRA-REF Field Organization.}~(top) Each experimental cultivar was planted in two different plots at distant locations in the field (borders were planted with well-known cultivars). In the top figure, each plot is labeled by its cultivar name, and plots from the same cultivar have matching colors. For each non-border cultivar, we include images from one of the plots in our training data, and images from the other plot in our test data, requiring models to generalize across field locations (as opposed to, for example, overfitting on unique ground features that are not relevant to the cultivar).~(bottom) Original data from the sensor is pre-processed to crop regions that confidently only consist of plants from a single plot. The blue rectangle in the image above shows a the ground boundaries of a plot projected onto the image.}
    \label{fig:field_map}
\end{figure}

The dataset is divided into a training dataset and a testing dataset. Each cultivar was grown in two separate plots in the TERRA-REF field as shown in~Figure~\ref{fig:field_map}~(top) to account for extremely local field or soil conditions that might impact the growth of plants in one particular plot. We leverage this natural split in the data when dividing our dataset between train and test -- images for a given cultivar in the training dataset come from one plot, while the test images from that same cultivar come from the other plot. This means that a model cannot achieve high performance by memorizing features that aren't meaningful phenotypes (e.g., by memorizing patterns observed in the dirt).

The original data from the TERRA-REF gantry is not divided by plot. In Figure~\ref{fig:field_map}~(bottom), we show an original RGB image captured by the TERRA-REF system, and one plot crop that was extracted from that image highlighted in blue. Plot crops were extracted based on the highly accurate metadata reported by the gantry regarding where the camera was positioned when the image was captured, and the known geographic boundaries of the plots on the ground. By defining the plot based on the ground coordinates, some parts of the plants can actually be cut off at the edges of the plot crop, due to the growth of the plants and the viewpoint of the sensor. We chose this behavior over the alternate of allowing a padding around each plot -- the padding approach would perhaps avoid the issue of chopping off portions of a plant in the images, but would introduce the potentially more problematic issue of having a significant number of pixels from adjacent plots (which come from different cultivars) included in the image.

\begin{figure*}[ht!]
    \centering
    \includegraphics[width=\textwidth]{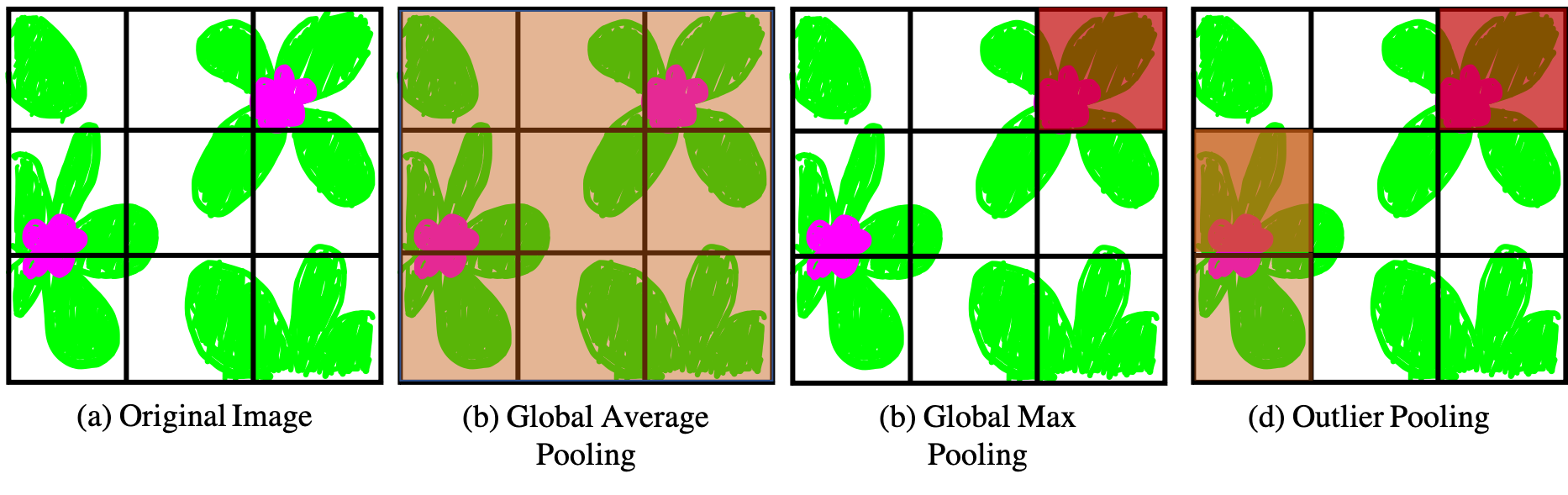}
    \caption{Imagine a hypothetical learned high level filter that activates highly at locations with pink flowers, like those shown in (a). In this figure, we visualize the activation of such a filter when passed through a Global Average Pooling layer (b), a Global Max Pooling layer(c), and our proposed Outlier Pooling (d). Higher activations being passed through the global pooling operation are visualized with darker red colors, while lower values are visualized in lighter colors (locations which do not contribute any activation have no coloring). In Global Average Pooling, the filter averages together active and in-active regions, under-weighting the flowers in the image. Global Max Pooling only considers one of the flowers. Our proposed Outlier Pooling passes activations for regions near both flowers.}
    \label{fig:pooling}
\end{figure*}

\section{Global Pooling Strategies}
A very common template for deep convolutional neural network architectures is to end with a final convolutional layer that is significantly smaller spatially than the input volume with a large number of channels, followed by a `global pooling' operation, which converts each channel to a single number. This global pooled output can then be passed through one or more fully connected layers. This template is seen in a variety of popular models used in both classification and deep metric learning, including AlexNet~\cite{krizhevsky2012imagenet}, ResNet~\cite{resnet}, ResNeXt~\cite{resnext}, Inception~\cite{inception} and DenseNet~\cite{densenet}.

The two most common strategies for performing the global pooling that converts each channel in the final convolutional layer to a single number are Global Average Pooling, where the value in the pooled layer, $n$, per channel, $j$, is the average value of that channel in the preceding layer:

\begin{equation}
\textup{layer}_n(j) = \frac{\sum{\textup{layer}_{n-1}(j)}}{\textup{width}_{n-1} \times \textup{height}_{n-1}}
\end{equation}

and Global Max Pooling, where the output at channel $i$ is determined by the max value of channel $j$ in the preceding layer:

\begin{equation}
\textup{layer}_n(j) = \max{(\textup{layer}_{n-1}(j))}
\end{equation}

Both approaches were introduced in~\cite{lin2013network}. While there are other global pooling strategies, such as global sum pooling (which takes the sum of all of the activations, and is proportional to global average pooling) and learned global pooling operators~\cite{ZHANG201836}, global average pooling and global max pooling are by far the most commonly used.

Intuitively, we can reason that a Global Average Pooling strategy is well suited to problem domains where the relevant features are evenly distributed throughout an image, while Global Max Pooling is well suited to problem domains where the relevant features are sparse. In high throughput phenotyping (especially in this gantry based system), however, relevant features are often sparse, such as panicles or flowers that are only present in some images from some species from some parts of the season, but may have more than one instance where present (such as the pink flowers in Figure~\ref{fig:pooling}). In these cases, global average pooling would under-weight the multiple instances of the interesting object, while global max pooling would only consider one of them.

\subsection{Outlier Pooling}
We propose a novel global pooling strategy that is well suited to these settings where relevant features may be interspersed throughout an image. The approach is called Outlier Pooling, where the output at channel $j$ is determined by the average value of any locations in the preceding layer that exceed a threshold:

\begin{equation}
\textup{layer}_n(j) = \frac{\textbf{I}\sum{\textup{layer}_{n-1}(j)}}{\sum{\mathbf{I}}}
\end{equation}

where $\mathbf{I}$ is an indicator function for every spatial location, $x$, in channel $j$ of $\textup{layer}_{n-1}$:

\begin{equation}
    \mathbf {I} _{x_j}:= \begin{cases}1~&{\text{ if }}~x_j >= t_j\\0~&{\text{ if }}~x_j < t_j\end{cases}
    \label{eq:indicator}
\end{equation}

and $t_j$ is the threshold for channel $j$:

\begin{equation}
    t_j = \mu_{\text{layer}_{n-1}(j)} + \lambda\sigma_{\text{layer}_{n-1}(j)}
\end{equation}

$\mu_{\text{layer}_{n-1}(j)}$ is the channel mean computed per image, $\sigma_{\text{layer}_{n-1}(j)}$ is the channel standard deviation computed per image, and $\lambda$ is a parameter that is fine-tuned based on the dataset (we empirically find a value of 2 to perform best for the Sorghum-100 dataset).

\subsection{Dynamic Outlier Pooling}\label{subsec:dynamic}
Very early in the training process, the most relevant features have not yet been learned. When initializing features from a model pretrained on ImageNet, all of the images in the Sorghum-100 dataset produce extremely similar output features as there is very low variability in these images relative to the entire space of ImageNet images. This means that there are very few locations whose activations will vary significantly from the mean, leading to difficulty in training with Outlier Pooling early on.

To account for this challenge, we use a Dynamic Outlier Pooling approach, which acts more like average pooling early in the training process and becomes progressively more focused on outliers as the training progresses. In this Dynamic Outlier Pooling, the output for a channel $j$ is:

\begin{equation}
    \textup{layer}_n(j) = \frac{w_1\textbf{I}\sum{\textup{layer}_{n-1}(j)} + w_2 (1-\textbf{I})\sum{\textup{layer}_{n-1}(j)}}{\textup{width}_{n-1} \times \textup{height}_{n-1}}
\end{equation}

where $\textbf{I}$ is the indicator function from Equation~\ref{eq:indicator}, and $w_1$ (the relative weighting of the values above the threshold) and $w_2$ (the relative weighting of the values below the threshold) change over the course of training according to the following schedule:

\begin{equation}
    w_1=1+\frac{\text{current epoch}}{\text{total epoches}}
\end{equation}
\begin{equation}
    w_2=1-\frac{\text{current epoch}}{\text{total epoches}}
\end{equation}

\begin{figure}
    \centering
    \includegraphics[width=\columnwidth]{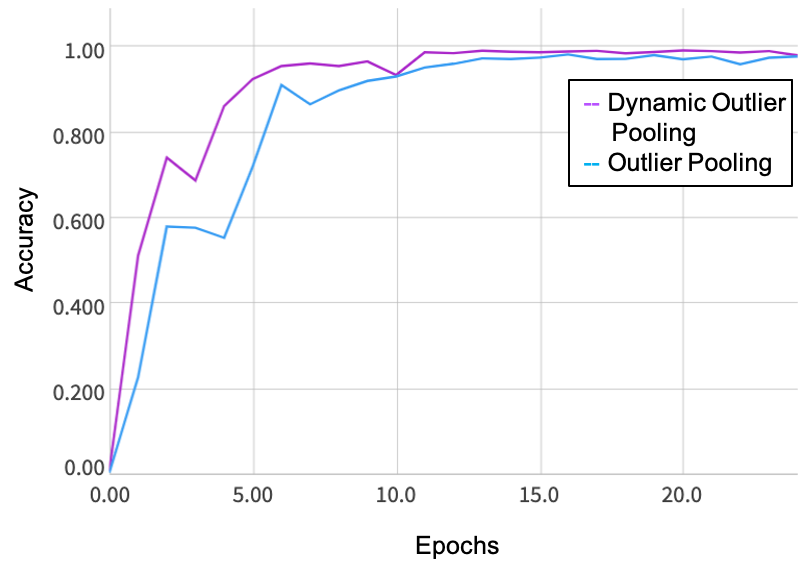}
    \caption{Using Dynamic Outlier Pooling, as described in Section~\ref{subsec:dynamic} leads to faster convergence on the Sorghum-100 training set when compared with the non-Dynamic Outlier Pooling.}
    \label{fig:dynamic_vs_outlier}
\end{figure}

The impact of this dynamic weighting is that as training progresses, the features that do not exceed the threshold become less impactful on the global pooled feature in $\text{layer}_n$. A comparison of the convergence between Dynamic Outlier Pooling and the non-Dynamic version is shown in Figure~\ref{fig:dynamic_vs_outlier}, where we can see that the Dynamic version leads to faster convergence.

\begin{figure*}
    \centering
    \includegraphics[width=\textwidth]{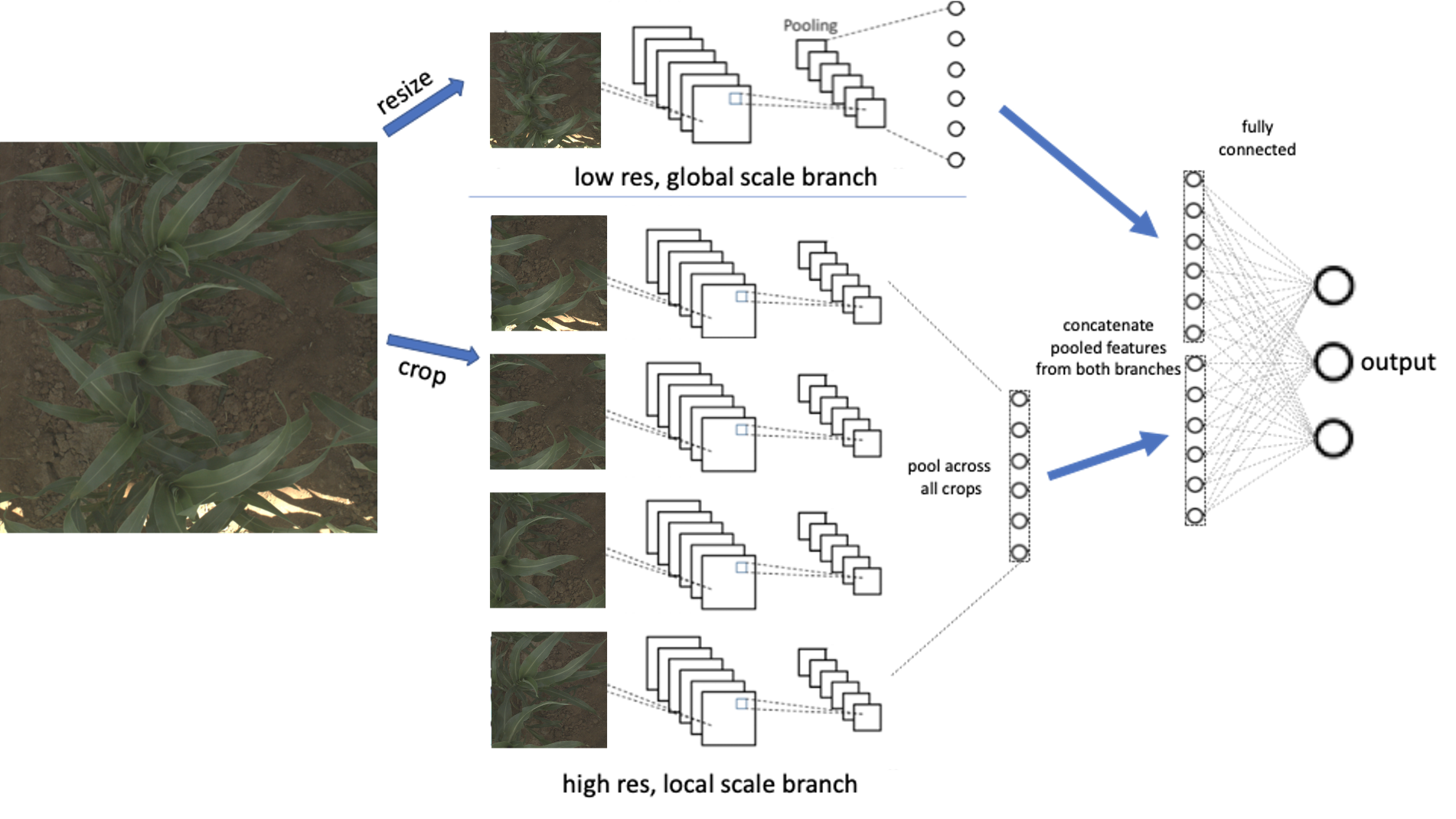}
    \caption{We propose a parallel network architecture that consists of both a low resolution, global scale branch, and a high resolution, local scale branch. While this diagram only shows a single convolutional layer followed by a global pooling operation, in reality we use a ResNet-50 trunk. In the high resolution branch, we take four random crops at original resolution, and perform global pooling across all four crops' final convolutional layers. The features from the two branches are then concatenated and fed through a final fully connected layer.}
    \label{fig:architecture}
\end{figure*}

\subsection{Multi-Resolution Crop Pooling}
In addition to considering the appropriate pooling strategy for the gantry-based high throughput phenotyping domain, it is also important to consider the best visual scale of data to consider. The recent advances in commercial GPU technologies with large amounts of memory (e.g., 16GB Nvidia V100 GPUs, such as those used in the experiments in this paper) allow for training on relatively high resolution imagery -- but there are still limits to the practical image resolution and size that can be used in training, and trade-offs between the size of the image and the number of images that can be fit in a batch. Moreover, it is desirable that the neural network `see' both global scale features (e.g., ``What does this whole plot of plants look like?'') as well as local, high resolution features (e.g., ``What does the structure of this flower look like?''). In order to do this while still having reasonable batch sizes and training, we propose a Multi-Resolution Crop Pooling approach, as seen in Figure~\ref{fig:architecture}.

This architecture has two parallel branches -- one that is a low resolution `global scale' branch, and one that is a high resolution `local scale' branch. In the global scale branch, the input is a resized version of the original image (during training, the image is resized and then randomly cropped to be square; during testing, the image is resized and then center cropped).

In the local scale branch, there are four models with shared weights. A random crop at original resolution is fed into each model. We then perform global pooling across the final convolutional layer from all four crops. The intuition behind this is that, as an example, you may see a flower in only one of the high resolution crops, and we want the network to give `full credit' for the presence of that flower, without down-weighting the features corresponding to that flower based on their absence in the other high resolution crops.

The global pooled features from these two branches are then concatenated and fed through a final fully connected layer.

This architecture is most similar to the feature pyramid architecture from~\cite{scales_cvpr}, which trained two networks: one on crops from the ImageNet~\cite{ILSVRC15} dataset for `object scale' features, and one on crops from the Places~\cite{zhou2014places} dataset, for `scene scale' features. At test time, they took different scale crops of the test image, ran each through whichever network the crop was suited for (smaller crops through the ImageNet trained network and larger crops through the Places trained network), computed the max feature per crop, and then concatened each of these max pooled features. Our approach differs in that we do not train our two networks on separate datasets, but rather different scales drawn from the same data, and most importantly in the cross-crop pooling that occurs before the features are concatenated with each other. We additionally utilize the proposed Dynamic Outlier pooling as the global pooling strategy both for the low resolution images in the global scale branch, and for the combination of the high resolution crops in the local scale branch.

\section{Experiments}
In the following experiments, we train separate networks comparing our Dynamic Outlier Pooling with Global Average Pooling and Global Max Pooling. We additionally do this comparison training a single low resolution image of the entire plot (only the global scale branch of our proposed network), four high resolution crops (only the high res branch of our proposed network), and finally our proposed multi-resolution model.

We initialize the trunk of each of our models with a ResNet-50 model~\cite{resnet} pre-trained on ImageNet. For experiments where only a single, low resolution image is used, during training we resize the original images to be $512$ pixels on its shortest side, and then take a random $512 \times 512$ crop (at test time, we take a center crop). For experiments where high resolution crops are used, we first resize the image so its shortest side is $1024$ pixels, and then take four random $512 \times 512$ crops (at test time, we take the four non-overlapping crops from a $1024 \times 1024$ center crop).

We normalize by channel means and standard deviations and perform random horizontal and vertical flips. For experiments using dynamic outlier pooling, we experimentally found a outlier pooling threshold of $2.0$ yielded the best performance on this dataset.

\begin{table*}[ht!]
    \centering
    \begin{tabular}{cccc}
    \toprule
    \diagbox{Crops (Resolution)}{Pooling} & Average & Max & Dynamic Outlier  \\
    \midrule
    Whole Image (low res) & 0.72137 & 0.64615 & 0.73829 \\
    Multi-crop (just high res) & 0.74296 & 0.6518 & 0.76923 \\
    Multi-crop (high and low res) & 0.76326 & 0.65486 & \textbf{0.78788} \\
    \bottomrule
    \end{tabular}
    \caption{In this table, we show the accuracy of different cropping and global pooling strategies on the Sorghum-100 test set. The performance of using high resolution crops is higher than using only one global, low resolution crop, but using both high and low resolution crops performs either individually. Additionally, using the Dynamic Outlier Pooling approach out-performs either Global Average Pooling or Global Max Pooling regardless of the resolution considered.}
    \label{tab:results}
\end{table*}

In Table~\ref{tab:results}, we report the results of an ablation study of the different combinations of image resolutions, as well as different global pooling strategies. We tune the learning rate for each experiment, and report results after 20 epochs. The results show that using multiple high resolution crops, regardless of the global pooling strategy, outperforms using the more global low resolution crop of the whole plot, and that combining the high resolution and low resolution images outperforms training just the global or local scale networks individually. Additionally, we can see that our proposed dynamic outlier pooling approach significantly outperforms both global max pooling and global average pooling, regardless of the resolutions considered.


\section{Conclusion}
In this paper, we introduced the Sorghum-100 cultivar classification dataset which includes tens of thousands of images from 100 bio-energy lines of sorghum grown in the TERRA-REF field. This gantry-based sorghum dataset for sorghum cultivar classification has higher temporal and spatial resolution than similar UAV based datasets, making the data suitable for generating models for true high-throughput phenotyping in the field. While this paper presents the dataset in the context of cultivar classification, we hope to in the future release more data products and metadata, including data from additional sensors and both hand- and algorithm- generated phenotypes with the goal of supporting the development of machine learning models for more sophisticated high-throughput phenotyping tasks.

In addition, we presented a novel approach to training a deep convolutional neural network on the cultivar classification task, using a multi-resolution network architecture, and new global average pooling strategy called outlier pooling that is well suited to agricultural domains where the relevant features may be interspersed sparsely throughout the imagery. We performed an ablation highlighting the improvement in the classification task based on both the multi-resolution network architecture and the new outlier pooling approach, and show that the best performance is achieved when the two components are combined.

Possible modifications to the multi-resolution outlier pooling approach include having a larger discrepancy between the resolutions seen in the global and local scale branches, considering a feature pyramid approach that is more complex than simply two different resolutions, computing channel thresholds per batch rather than per image, and allowing varying threshold multipliers for different channels. While we look forward to exploring these questions in future work, the simple settings chosen in this work achieve significantly superior performance over baselines that consider only a single resolution or standard global pooling approaches.

{
\bibliographystyle{ieee_fullname}
\bibliography{cvpr}
}

\end{document}